\documentclass[runningheads]{llncs}
\usepackage[T1]{fontenc}

\usepackage{amsmath}
\usepackage{hyperref}       
\usepackage{url}            
\usepackage{booktabs}       
\usepackage{graphicx}
\usepackage{nicefrac}       
\usepackage{microtype}      
\usepackage{xcolor}         
\usepackage{multirow}
\usepackage{makecell}
\usepackage{subfigure}
\usepackage{graphicx}
\usepackage{amssymb}
\begin{document}
\title{Latent Neural Operator Pretraining for Solving Time-Dependent PDEs}
%
%
\author{Tian Wang\inst{1,2} \and Chuang Wang\inst{1,2,\ast}}
\authorrunning{}
%
\institute{Institute of Automation, Chinese Academy of Sciences, Beijing 100190, China
\email{\{wangtian2022,wangchuang\}@ia.ac.cn}\\ \and
School of Artificial Intelligence, University of Chinese Academy of Sciences, Beijing 100049, China\\
}
\include{command}
\global\long\def\tw#1{{\color{blue} Tian: #1}}
\global\long\def\cw#1{{\color{red} Chuang: #1}}

\maketitle              

\begin{abstract}
\footnotetext[0]{$\ast$ Corresponding Author. This work is supported by Beijing Municipal Science and Technology Project (No. Z231100010323005), and  Pioneer Hundred Talents Program of CAS  under Grant Y9S9MS08 and E2S40101.} 
Pretraining methods gain increasing attraction recently for solving PDEs with neural operators. It alleviates the data scarcity problem encountered by neural operator learning when solving single PDE via training on large-scale datasets consisting of various PDEs and utilizing shared patterns among different PDEs to improve the solution precision. In this work, we propose the Latent Neural Operator Pretraining (LNOP) framework based on the Latent Neural Operator (LNO) backbone. We achieve universal transformation through pretraining on hybrid time-dependent PDE dataset to extract representations of different physical systems and solve various time-dependent PDEs in the latent space through finetuning on single PDE dataset. Our proposed LNOP framework reduces the solution error by 31.7\% on four problems and can be further improved to 57.1\% after finetuning. On out-of-distribution dataset, our LNOP model achieves roughly 50\% lower error and 3$\times$ data efficiency on average across different dataset sizes. These results show that our method is more competitive in terms of solution precision, transfer capability and data efficiency compared to non-pretrained neural operators. 

\end{abstract}

\section{Introduction}
\vspace{-5pt}
Partial Differential Equations (PDEs) describe the underlying principles of numerous phenomena in the real world with broad applications ranging from weather forecasting\cite{pangu-weather,Fourcastnet} and pollution detection\cite{FINN_GroundwaterContaminant} to industrial designing\cite{ComplexStress}. Consequently, solving PDE accurately and efficiently remains a pivotal research area. Traditional numerical methods such as finite element method and spectral method solve PDEs 
by transforming continuous differential equations into discrete difference equations, which requires specialized knowledge and substantial computational resources. With the advent of deep learning, employing surrogate models based on neural network as alternatives to traditional numerical methods offers a fresh approach to PDE solving with cheaper computational requirement and potentially better generalizability to characterize the dynamics of the real world even beyond the scope of PDE-based models.

Neural operator\cite{NO}, which adopts the data-driven paradigm to directly learn infinite-dimensional mappings from input functions to output functions, stands out as a promising surrogate modeling approach for solving PDEs. Compared to its peer methods, {\em e.g.}, physics-informed numerical networks\cite{PINN,FINN,PeRCNN}, neural operators demonstrate faster inference speeds and superior generalization ability but less accuracy. Existing neural operator methods mostly rely on training different models for corresponding PDE problems based on simulated data generated by numerical methods,  restricting neural operator to solve specialized, case-by-case problems, rather than leveraging extensive data from diverse PDE problems to cover common representations for solving general problems.

Pretraining framework has become a {\em de facto} groundwork for building universally capable large models in computer vision \cite{SimCLR,MoCo,CLIP} and natural language processing\cite{LDM,GPT3}. In the realm of PDE solving, pretraining promises a prospective solution to alleviate the data-scarcity problems for  neural  operator learning.

In this work, we follow the idea of Latent Neural Operator\cite{LNO} (LNO) and establish pretraining for multiple time-dependent PDEs in the latent space. The major contributions are summarized as follows.
\begin{itemize}
\item We introduce the Latent Neural Operator Pretraining (LNOP) framework of extracting representations of different PDEs in a shared latent space through a universal transformation module implemented as Physics-Cross-Attention (PhCA). 
\item Numerical experiments on various timed-dependent PDE problems indicate that our proposed framework exploits common representations from various physical systems and achieves better solution precision compared to methods without pretraining.
\item The proposed pretrain framework also exhibits strong transfer learning capability and data efficiency with the pretrained universal transformation tested on experiments of out-of-distribution time-dependent PDE problems.
\end{itemize}

\vspace{-10pt}
\paragraph{} The rest of the paper is organized as follows. Firstly, we introduce existing works of neural operator and PDE pretraining in Section \ref{related-work}. We illustrate the working principles of LNO backbone and our proposed LNOP framework in Section \ref{method}. We describe the dataset details in Section \ref{dataset}. Subsequently, we present a series of experiments and corresponding analysis to study the performance of our LNOP framework in terms of solution precision, transfer capability and data efficiency in Section \ref{experiment}. Finally, we provide concluding remarks of our work in Section \ref{conclusion}.

\vspace{-5pt}
\section{Related Work}
\label{related-work}
\vspace{-5pt}
\subsection{Neural Operator}
\vspace{-5pt}
Neural operator methods aim to solve PDEs by learning mappings between functions. For instance, they map the coefficients, initial conditions or boundary conditions of PDEs, which serve as input functions, to the solution, which serves as the output function. DeepONet\cite{DeepONet} designs trunk and branch structures for encoding query positions of the output function and observed values of the input function respectively, and the results from these two parts are combined to predict the output function. FNO\cite{FNO} utilizes Fourier transform to learn transformations between functions in the frequency domain and derives a series of variants including Geo-FNO\cite{GeoFNO}, U-FNO\cite{UFNO}, and F-FNO\cite{FFNO}, which extend the applicability or enhance the precision and efficiency of FNO.

With the tremendous success of Transformer\cite{Attention} structures in fields of computer vision and natural language processing, Transformer-based neural operator methods have also been proposed. Galerkin-Transformer\cite{GalerkinTransformer} firstly introduces Galerkin-type and Fourier-type attention mechanisms as kernels in neural operator. OFormer\cite{OFormer} extends Galerkin-type attention to the case of cross-attention. GNOT\cite{GNOT} further proposes Heterogeneous Normalized Cross-Attention to accommodate multiple input functions. 

To address the issues of the significant computational cost of quadratic complexity attention mechanisms when applied to PDE problems with large spatial grids, many works have been proposed. FactFormer\cite{Factformer} projects high-dimensional PDEs into multiple single-dimensional functions. Transolver\cite{Transolver} uses physical attention to allocate geometric features to a constant number of physical slices in each Transformer block. LNO\cite{LNO} employs Physics-Cross-Attention (PhCA) to solve PDEs in the latent space. Our LNOP framework follows the idea of LNO, where we train encoder and decoder to learn universal transformation which extracts the common representations of multiple PDEs in a shared latent space.

Despite the strong nonlinear approximating capability of neural operator, it faces the challenge of insufficient training data. Simulated data generated using traditional numerical methods is often computationally expensive, while real-world data is difficult to collect. Some approaches such as PINO\cite{PINO} and PI-DeepONet\cite{PI-DeepONet} incorporate physical priors into neural operator to alleviate data scarcity. Other works attempt to construct pretrained foundation models for various downstream PDE tasks involving scarce data. 

\vspace{-5pt}
\subsection{PDE Pretraining}
Pretraining has been a crucial driver of recent breakthroughs in fields of computer vision\cite{MAE,BLIP2} and natural language processing\cite{Bert,LLaMa,GPT3}. Models are firstly pretrained on pretext tasks to learn generic and task-agnostic representations from vast amounts of raw data in an unsupervised manner. The task-specific components are then finetuned with minimal data to complete the downstream task. This paradigm significantly enhances data efficiency and enables effective utilization of resources.

In the field of PDE solving, pretraining strategies are also gradually being adopted. MPP\cite{MPP} firstly proposes autoregressive task to pretrain video Transformer\cite{ViViT} models on dataset consisted of various time-dependent PDEs in fluid mechanics. DPOT\cite{DPOT} extends MPP by incorporating denoising objective in autoregressive tasks and utilizes a Fourier Transformer\cite{AFNO,FNet} backbone. PDEformer\cite{PDEFormer} trains graph Transformers on dataset containing 1D time-dependent PDEs with diverse conditions. Unisolver\cite{Unisolver} extends the idea of equation conditional modulation to the case of 2D time-dependent PDEs and applies domain-wise and point-wise conditions to modulate PDE representations using different attention heads. Both of these efforts and our proposed framework follow the pretraining and finetuning paradigm same as in fields of computer vision and natural language processing.

\vspace{-5pt}
\section{Method}
\vspace{-5pt}
\label{method}
We first provide the formal definition of solving time-dependent PDEs, and then introduce the Latent Neural Operator (LNO) backbone. Finally, we present the framework of our Latent Neural Operator Pretraining (LNOP) approach.

\vspace{-5pt}
\subsection{Problem Setup}
We consider the time-dependent PDE defined on $D \subseteq \Omega \times [0,T]$
\vspace{-5pt}
\begin{equation}\label{pde_def}
  \begin{aligned}\nonumber
  \mathcal{L}_{a}\circ u=0, \quad \text{with }
    u({x,0})=u_0({x}), \;x \in \Omega, \quad \text{and }
    u({x,t})=b({x}),\; x\in \partial \Omega, 
  \end{aligned}
  \vspace{-5pt}
\end{equation}
where $\mathcal{L}_{a}$ is an operator containing partial differentials parameterized by coefficients $a$;  $u_0(x)$ and $b(x)$ represent the initial condition and boundary condition respectively, and $u(x,t)=u_t(x)$ is the solution to the PDE. 

A set of classic linear time-dependent PDEs such as the heat equation, Laplace's equation and the wave equation can be decoupled and  solved by a proper functional transformation, following a general procedure that i) first transform the spatial domain of the system into another new domain; ii) then predict the evolution of the system over time in the new domain; iii) finally transform back from the new domain to the spatial domain. This method can be formalized as
\vspace{-5pt}
\begin{equation}
\begin{aligned} \nonumber
    \hat{u}_0(w)=\mathcal{F}(u_0)(x),\quad \hat{u}_{t+\Delta t}=\mathcal{P}(\hat{u}_t),\quad u_T(x)=\mathcal{F}{^{-1}}(\hat{u}_T)(w), \\
  \end{aligned}
  \vspace{-5pt}
\end{equation}
where $x\in \Omega, w \in \Omega'$. For different time-dependent PDE systems, the transformation $\mathcal{F}$ can be, for example, the Fourier transform or Laplace transform. Inspired by this, we treat the PhCA encoder and decoder modules in LNO\cite{LNO} as universal feature transformation and its inverse respectively. We consider the intermediate Transformer layers as short-term propagator, and train the LNO end-to-end to learn feature transformation that is applicable across different PDE problems.

\vspace{-10pt}
\subsection{Latent Neural Operator}
\label{lno}
We briefly introduce our previous work of LNO\cite{LNO} for solving forward and inverse PDE problems as the pretraining backbone. Contrast to the original work\cite{LNO} that trains models to predict different PDEs separately in a case-by-case manner, we explore the pretraining and universal representation ability of LNO in this work.

LNO consists of five modules: an input projector to lift the dimension of the input data, an encoder to transform the input embedding into a learnable latent space, a sequence of Transformer layers for modeling operator in the latent space, a decoder to recover the latent representation back to the real-world space and an output projector to project back the lifted dimension of the output data.

Physics-Cross-Attention (PhCA) is the core of LNO, used for transforming between $N$ embeddings in the real geometric space and $M$ representation tokens in the latent space. Since the latent space is much more compact than the large geometric space, PhCA significantly reduces the computational load when solving PDEs.

The input projector contains two parts, a branch projector and a trunk projector, following the convention of DeepONet \cite{DeepONet}. In the encoding phase: i) the branch projector converts $N$ observation positions and corresponding values of the input function into $N$ embeddings in real geometric space, which serve as the value matrix; ii) the trunk projector lifts the dimension of each observation position of the input function, which is then converted into $M$ attention scores through MLP in the PhCA encoder, and $N$ observation positions yield $M \times N$ attention score matrix; iii) the row-normalized attention score matrix is multiplied by the value matrix to obtain the representation tokens in the latent space.

Conversely, the decoding phase is a inverse process of the encoding. Specifically, i) the representation tokens transformed through multiple Transformer layers serve as the value matrix; ii) the trunk projector lifts the dimension of each query position of the output function, which is then converted into $M$ attention scores through MLP in the PhCA decoder, and $N$ query positions yield $M \times N$ attention score matrix;  iii) the column-normalized attention score matrix is transposed and multiplied by the value matrix to map the representation tokens in the latent space back to the real geometric space, which finally converted into output function values by the output projector.

\vspace{-10pt}
\subsection{Frame Design}
\label{frame}
\paragraph{Overall} Our proposed LNOP framework is an end-to-end method for solving multiple time-dependent PDEs based on LNO backbone. Unlike the original LNO work, we adopt a hybrid pretraining strategy to guide the PhCA mechanism in learning the universal transformation applicable to different heterogeneous physical systems. This extends the LNO, which was designed to solve a single physical system, into the LNOP capable of solving multiple physical systems in higher precision. 

\begin{figure}[htbp]
\centering
\includegraphics[width=1.0\textwidth]{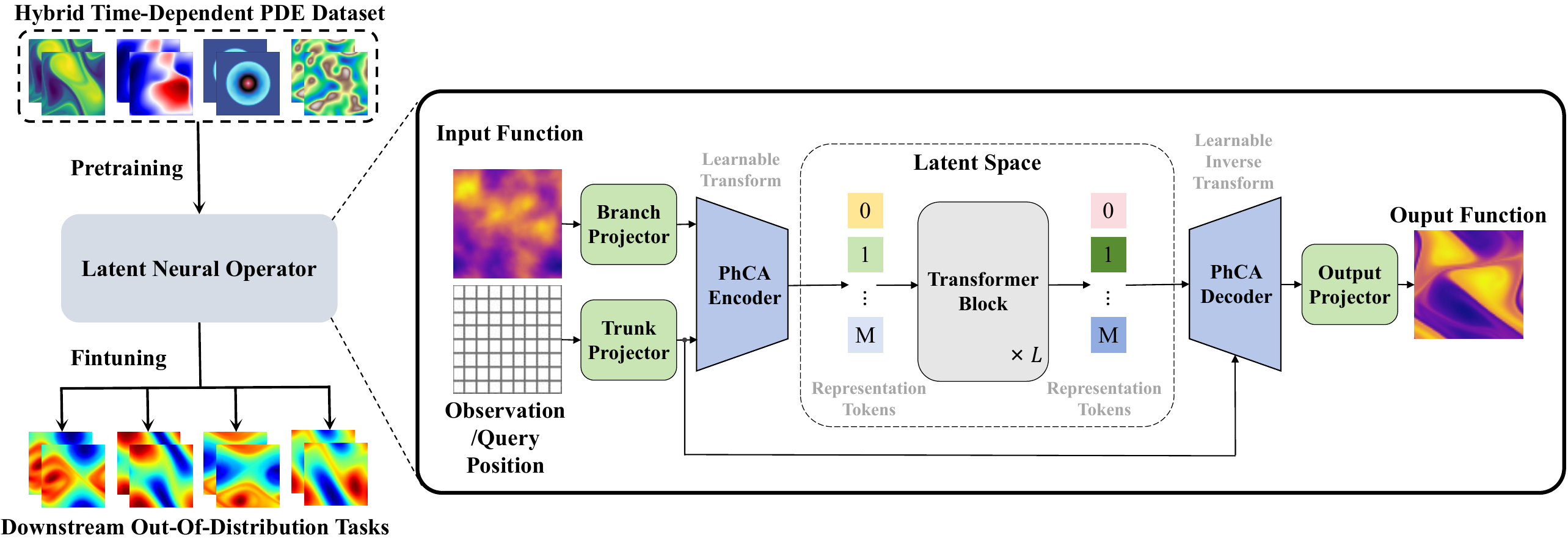}
\caption{The overall architecture of Latent Neural Operator Pretraining. We pretrain the Latent Neural Operator (LNO) on datasets containing various time-dependent PDE problems. This enables the PhCA encoder and decoder within the LNO to learn general transformation, which is used to extract PDE representations in the latent space. Subsequently, we finetune the LNO on out-of-distribution downstream tasks to apply the learned universal transformation.}
\label{arch}
\end{figure}

LNOP consists of three components: PhCA encoder/decoder and the propagator. As illustrated in Figure \ref{arch}, we pretrain the LNO backbone on the hybrid dataset consisting of various time-dependent PDEs, and then finetune it on downstream tasks for solving other PDEs. 

\paragraph{Pretraining PhCA Encoder/Decoder} In this work, we explore the ability of PhCA from efficiency in \cite{LNO} to universality. Unlike the LNO method where multiple models are required to construct separated latent spaces for different PDE problems, the LNOP framework uses a single model to construct the shared latent space for various PDE problems during pretraining. In this context, the PhCA encoder/decoder can be seen as learnable universal transformation. The PhCA encoder transforms the sampled functions in the real geometric space into representations in the latent space, while the PhCA decoder reconstructs the function information from these representations. PhCA encoder and decoder operate on the PDE spatial states and do not involve the temporal evolution process. 

The PhCA encoder and decoder compress high-dimensional PDE representations in the raw real-world space from single physical system into representations that retain the essence of the PDE spatial states with higher information density in the latent space. This enhances the efficiency of simulating interactions between different spatial locations of the PDEs. In LNOP, the objective of solving multiple physical systems simultaneously in hybrid dataset further constrains the universal transformation learned by the PhCA encoder and decoder. As a result, the representations capture the commonalities of different PDE spatial states, facilitating their transfer to other downstream tasks.

\paragraph{Finetuning Propagator} Since solving time-dependent PDEs essentially involves calculating the system's response based on its spatial state at current moment and its inherent spatial interaction rules, we need the propagator to compute the representations of the spatial state at next moment based on the representations extracted by the PhCA encoder/decoder from the current PDE spatial state. 

Different physical systems have distinct spatial interaction rules, so different propagators are needed for each physical system in principle. However, during pretraining process, we use only one propagator to help align the representation space derived from PhCA encoder/decoder with the solution space used for PDE temporal evolution prediction. This propagator predicts the new state of the PDE system at next time step in each forward call of the entire model to reduce the approximating errors caused by differences among various physical systems. This means that each short-time evolution prediction of the PDE goes through the complete process involving the PhCA encoder/decoder and the propagator. During finetuning process, we adjust the propagator's parameters for different PDE problems to approximate the corresponding spatial interaction rules.

Considering the attention mechanism's ability to model interactions among multiple vectors and its excellent performance as a kernel function in operator learning, we choose to use a series of Transformer layers as the propagator.

\paragraph{} In the above proposed LNOP framework, we pretrain the LNO backbone on hybrid dataset containing multiple time-dependent PDEs, rather than training separate LNO model for each PDE as done in the original LNO paper\cite{LNO}. Throughout this pretraining process, we expect that the PhCA encoder/decoder can learn a universal transformation  which maps spatial domain features of different PDEs to a shared latent space, so that the propagator can perform evolution prediction in the temporal domain within this latent space. Details on training procedure and comparison are presented in the experiment section.

\vspace{-5pt}
\section{Dataset}
\vspace{-5pt}
\label{dataset}
We consider the hybrid dataset containing multiple physical systems which are all time-dependent PDEs in 2D space, including the Navier-Stokes equation, Shallow-Water equation, Burgers' equation and Reaction-Diffusion equation. All these PDEs describe time-varying systems whose response is determined by interactions among different spatial locations. Therefore, we can solve them by neural network which extracts representations of PDE spatial states and approximates the temporal evolution of the representations.
\vspace{-5pt}
\paragraph{Navier-Stokes Equation} We use the Navier-Stokes equation in the FNO dataset\cite{FNO} with the form of
\vspace{-6pt}
\begin{equation} \nonumber
    \begin{aligned}
        \partial_tu(x,t)+w(x,t)\cdot \nabla u(x,t)&=\upsilon \Delta u(x,t)+f(x) \\
        \nabla \cdot w(x,t)&=0 \\
         x\in \Omega, t&\in[0,T],
    \end{aligned}
    \vspace{-6pt}
\end{equation}
where $w$ is the velocity, $u=\nabla \times w$ is the vorticity, $\upsilon$ is the viscosity coefficient and $f(x)$ is the forcing term. This equation is the fundamental equation used to explain and predict the behavior of fluids under various conditions.

We set $\Omega=[0,1]^2, T=20 $ and $f(x)=0.1(sin(2\pi(x_1+x_2))+cos(2\pi(x_1+x_2)))$. The initial condition is generated according to $u(x,0) \sim \mathcal{N}(7^{2/3}(-\Delta + 49I)^{-2.5})$.  Periodic boundary condition is applied. We use the data under three different viscosity coefficient values $\upsilon=10^{-5},10^{-4},10^{-3}$. For $\upsilon=10^{-5}$, there are $1200$ trajectories each containing $20$ frames on $64\times 64$ spatial grids. We use $1100$ trajectories for training and the rest $100$ for testing. For $\upsilon=10^{-4}$ and $\upsilon=10^{-3}$, there are $1100$ trajectories respectively, each containing $25$ frames on $64\times 64$ spatial grids. We use $1000$ trajectories for training and the rest $100$ for testing. The data with viscosity coefficient $\upsilon=10^{-5}$ will be used during pretraining, while the data with viscosity coefficients $\upsilon=10^{-4},10^{-3}$ will be used to evaluate the model's transfer capability.
\vspace{-5pt}
\paragraph{Shallow-Water Equation} We use the Shallow-Water equation in the PDEBench\cite{PDEBench} dataset which has the form of
\vspace{-3pt}
\begin{equation} \nonumber
    \begin{aligned}
        \partial_tu(x,t)+\partial_xu(x,t)v(x,t)&=0 \\
        \partial_tu(x,t)v(x,t)+\partial_x(u(x,t)v^2(x,t)+\frac{1}{2}g_ru^2(x,t))&=-g_ru(x,t)\partial_xb \\
        x\in\Omega,t\in[0,T],
    \end{aligned}
    \vspace{-3pt}
\end{equation}
where $u$ is the water depth, $v$ is the velocities in horizontal and vertical direction, $b$ is the spatially varying bathymetry and $g_r$ is the gravitational acceleration. This equation can be used to describe the fluid motion in shallow water regions.

We set $\Omega=[-2.5,2.5]^2, T=1$. The initial condition is
\vspace{-3pt}
\begin{equation} \nonumber
u(x,0)=\left\{
\begin{aligned}
2.0, \quad \Vert x \Vert^2_2 > r \\
1.0, \quad \Vert x \Vert^2_2 \leq r,
\end{aligned}
\right.
\vspace{-3pt}
\end{equation}
where $r\sim \mathcal{U}(0.3,0.7)$ is the radius of the circular bump in the center of the spatial domain. Dirichlet boundary condition is applied. There are $1000$ trajectories each containing $100$ frames on $128\times 128$ spatial grids. We downsample the temporal dimension to $20$ and the spatial dimensions to $64 \times 64$. We use $900$ trajectories for training and the rest $100$ for testing. Although the data we generated exhibits spatial symmetry, we do not incorporate this characteristic as a prior into the model architecture or predicting process.
\vspace{-5pt}
\paragraph{Burgers' Equation} We generate the data of Burgers'
equation following
\begin{gather*}
    \partial_tu(x,t)=D\partial_{xx}u(x,t)-u(x,t)\partial_xu(x,t) \\
    x\in\Omega,t\in[0,T],
\end{gather*}
where $D$ is the diffusion coefficient. This equation is used to simulate the formation of shock waves.

We set $\Omega=[-1,1]^2,T=1$ and $D=0.001/\pi$. The initial condition is generated according to $u(x,0) \sim \mathcal{N}(7^{2/3}(-\Delta + 49I)^{-2.5})$. Periodic boundary condition is applied. There are $1200$ trajectories each containing $20$ frames on $64\times 64$ spatial grids. We use $1100$ trajectories for training and the rest $100$ for testing. 
\vspace{-5pt}
\paragraph{Reaction-Diffusion Equation} We generate the data of Reaction-Diffusion equation following
\begin{equation} \nonumber
\begin{aligned}
    \partial_tu_1(x,t)&=u_1-u_1^3-k-u_2+D_1\Delta u_1(x,t) \\
    \partial_tu_2(x,t)&=u_1-u_2+D_2\Delta u_1(x,t) \\
    &x\in\Omega,t\in[0,T],
\end{aligned}
\end{equation}
where $D_1, D_2$ are the diffusion coefficients for the activator and inhibitor respectively. This equation can be used to describe and analyze the material diffusion and chemical reaction processes.

We set $\Omega=[-1,1]^2,T=10$ and $D_1=10^{-3}, D_2=5\times10^{-3},k=5\times 10^{-3}$. The initial condition $u(x,0)$ is generated according to $u(x,0) \sim \mathcal{N}(7^{2/3}(-\Delta + 49I)^{-2.5})$. Neumann boundary condition is applied. There are $1200$ trajectories each containing $20$ frames on $64\times 64$ spatial grids. We use $1100$ trajectories for training and the rest $100$ for testing.

\vspace{-5pt}
\section{Experiment}
\vspace{-5pt}
\label{experiment}
We conduct a series of experiments on our dataset and compare the results with both classical and newly proposed neural operator methods. We demonstrate our LNOP framework effectively improves the solution precision for time-dependent PDEs, exhibits strong transfer capability and data efficiency.

\paragraph{Baselines} We choose to compare our framework with FNO\cite{FNO}, Transolver\cite{Transolver} and LNO\cite{LNO}. FNO is classical neural operator method which stacks Fourier layers to learn mappings between functions. Leveraging the significant role of Fourier transform in time-frequency analysis, FNO has found wide application across various tasks\cite{SFNO,Fourcastnet,CO2}. Transolver is the latest SOTA neural operator method, which performs allocation and re-allocation between geometric space and physical slices in each Transformer layer to help exploit the physical interactions between different spatial regions. LNO has been introduced in Section \ref{lno}. All of FNO, Transolver and LNO are trained on each single PDE problem in our dataset.

\paragraph{Implementation} All models are trained for $500$ epoch using AdamW\cite{AdamW} optimizer and OneCycleLR\cite{OneCycleLR} scheduler with initial learning rate $0.001$. We choose relative L2 error as the loss function. For FNO, we set the mode number to $12$.
For Transolver, we set the slice number to $32$. We construct both small-scale and large-scale versions of LNO. The small-scale version (marked with the suffix -S) consists of $4$ Transformer layers and has $64$ representation tokens each of $128$ dimension. The large-scale version (marked with the suffix -L) consists of $8$ Transformer layers and has $256$ representation tokens each of $256$ dimension. All experiments are conducted on a single RTX 3090 GPU, with batch sizes adjusted from $4$ to $16$ based on the memory usage. The FNO and Transolver has about $0.9$ and $1.6$ million model parameters respectively, while the two scale versions of LNO have $0.8$ and $5.0$ million model parameters respectively.
\vspace{-5pt}
\subsection{Solution Precision}
We train models of different methods to autoregressively solve various time-dependent PDEs given the initial 10 time steps and compare the solution precision, as shown in Table \ref{tab-precision}. 

LNOP trained on the hybrid dataset achieves higher or comparable solution precision on various PDE problems, even without finetuning, compared to LNO trained on individual dataset. The small-scale version and large-scale version LNOP method reduces the error by an average of 5.6\% and 31.7\%, respectively, compared to LNO across all problems.

As the finetuning epochs on each PDE problem's dataset increase from $100$ to $500$, the solution precision of LNOP continues to improve. After finetuning for $500$ epoch, the error reduction of LNOP method in two scale versions further improves to 46.8\% and 57.1\% respectively. These results demonstrate that the universal transformation learned by LNOP during pretraining can extract common representations across different PDE problems, thereby effectively raising solution precision.

LNOP achieves more significant precision improvements through pretraining in the large-scale version than the small-scale version. This indicates that as the model's capacity to process data increases, the data from different PDE problems can complement each other. This further underscores the necessity of learning universal transformation from multiple physical systems. 

\vspace{-5pt}
\begin{table}[htbp]
\centering
\caption{The solution precision of different models on various PDE problems respectively. Relative L2 error is recorded. The best result in each group is in bold. Values in parentheses indicate the change in error relative to the LNO model of the same scale, where '-' denotes a reduction and '+' denotes an increase.}
\label{tab-precision}
\scalebox{0.85}{
\begin{tabular}{|c|c|c|c|c|}
\hline
Model&Navier-Stokes&Shallow-Water&Burgers'&Reaction-Diffusion\\
\hline
FNO\cite{FNO}&0.1214&0.0017&0.0174&0.0572\\
Transolver\cite{Transolver}&0.1012&0.0015&0.0193&0.0473 \\
\hline
LNO-S\cite{LNO}&0.0949&0.0013&0.0148&0.0467\\
LNOP-S(pretrain)&0.0730(-23.1\%)&0.0014(+7.7\%)&0.0153(+3.4\%)&0.0419(-10.3\%)\\
LNOP-S(finetune-100)&0.0664(-30.0\%)&0.0013(-0\%)&0.0130(-12.2\%)&0.0367(-21.4\%) \\
LNOP-L(finetune-500)&\textbf{0.0456(-52.0\%)}&\textbf{0.0005(-61.5\%)}&\textbf{0.0112(-24.3\%)}&\textbf{0.0236(-49.5\%)} \\
\hline
LNO-L\cite{LNO}&0.0845&0.0014&0.0037&0.0052\\
LNOP-L(pretrain)&0.0328(-61.2\%)&0.0010(-28.6\%)&0.0029(-21.6\%)&0.0044(-15.4\%)\\
LNOP-L(finetune-100)&0.0302(-64.3\%)&0.0005(-64.3\%)&0.0025(-32.4\%)&0.0040(-23.1\%) \\
LNOP-L(finetune-500)&\textbf{0.0269(-68.2\%)}&\textbf{0.0003(-78.6\%)}&\textbf{0.0021(-43.2\%)}&\textbf{0.0032(-38.5\%)} \\
\hline
\end{tabular}
}
\vspace{-25pt}
\end{table}

\subsection{Transfer Capability}
An important goal of pretraining is to allow models to learn general representations from large amounts of data and provide better parameter initialization for downstream tasks, resulting in improvements in data utilization compared to training from scratch where parameters are randomly initialized. We expect LNOP to have strong transfer capability to help improving solution precision and data efficiency in downstream tasks involving out-of-distribution PDEs which are unseen during the pretraining process.

We finetune the pretrained LNOP model on Navier-Stokes equations with viscosity coefficients of $10^{-4}$ and $10^{-3}$ (as opposed to $10^{-5}$ during pretraining process) for $500$ epochs using different proportion of the total data amount to validate their data efficiency, and compare the solution precision with the baselines which are trained from scratch.

The results in Table \ref{tab-transfer1}, \ref{tab-transfer2} show that pretrained LNOP achieves higher solution precision than the baselines when finetuned with varying proportions of data on Navier-Stokes equations with out-of-distribution viscosity coefficients. For viscosity coefficients of $10^{-4}$, the LNOP model in two scale versions reduces the average error by 38.7\% and 49.8\%, respectively, compared to LNO across different training data sizes. For viscosity coefficients of $10^{-3}$, the error reductions are 57.9\% and 59.6\% respectively. The LNOP model finetuned using only 30\% data can achieve higher solution precision than the LNO model trained from scratch using 100\% data. Roughly, the LNOP has 3x data efficiency compared to the LNO. 

The results demonstrate that the universal transformation learned from in-distribution time-dependent physical systems can be effectively adapted to out-of-distribution time-dependent PDE problems even under low-data situation, proving that our proposed LNOP framework possesses strong transfer learning capability and highly efficient data utilization.

\begin{table}[ht]
\centering
\caption{The solution precision of different models on Navier-Stokes equation with viscosity coefficient values $\upsilon=10^{-4}$ of varying dataset scales. Relative L2 error is recorded. The best result in each group is in bold. Values in parentheses indicate the change in error relative to the LNO model of the same scale, where '-' denotes a reduction and '+' denotes an increase.}
\label{tab-transfer1}
\scalebox{0.75}{
\begin{tabular}{|c|c|c|c|c|c|}
\hline
Model&10\% data&30\% data&50\% data&80\% data&100\% data\\
\hline
FNO\cite{FNO}&0.6618&0.6289&0.6327&0.6076&0.5632 \\
Transolver\cite{Transolver}&0.4142&0.3872&0.3003&0.2498&0.2217 \\
\hline
LNO-S\cite{LNO}&0.3072&0.2224&0.1931&0.1571&0.1548 \\
LNOP-S(finetune-500)&\textbf{0.2424(-21.1\%)}&\textbf{0.1460(-34.4\%)}&\textbf{0.1090(-43.6\%)}&\textbf{0.0835(-46.8\%)}&\textbf{0.0815(-47.4\%)} \\
\hline
LNO-L\cite{LNO}&0.2912&0.1964&0.1702&0.1402&0.1292 \\
LNOP-L(finetune-500)&\textbf{0.2259(-22.4\%)}&\textbf{0.1227(-37.5\%)}&\textbf{0.0826(-51.5\%)}&\textbf{0.0477(-66.0\%)}&\textbf{0.0367(-71.6\%)} \\
\hline
\end{tabular}
}
\vspace{-30pt}
\end{table}

\begin{table}[ht]
\centering
\caption{The solution precision of different models on Navier-Stokes equation with viscosity coefficient values $\upsilon=10^{-3}$ of varying dataset scales. Relative L2 error is recorded. The best result in each group is in bold. Values in parentheses indicate the change in error relative to the LNO model of the same scale, where '-' denotes a reduction and '+' denotes an increase.}
\label{tab-transfer2}
\scalebox{0.75}{
\begin{tabular}{|c|c|c|c|c|c|}
\hline
Model&10\% data&30\% data&50\% data&80\% data&100\% data\\
\hline
FNO\cite{FNO}&0.0203&0.0097&0.0053&0.0041&0.0038 \\
Transolver\cite{Transolver}&0.0354&0.0112&0.0054&0.0039&0.0036 \\
\hline
LNO-S\cite{LNO}&0.0218&0.0073&0.0046&0.0033&0.0031 \\
LNOP-S(finetune-500)&\textbf{0.0048(-78.0\%)}&\textbf{0.0026(-64.4\%)}&\textbf{0.0020(-56.5\%)}&\textbf{0.0018(-45.5\%)}&\textbf{0.0017(-45.2\%)} \\
\hline
LNO-L\cite{LNO}&0.0128&0.0036&0.0023&0.0019&0.0016 \\
LNOP-L(finetune-500)&\textbf{0.0024(-81.3\%)}&\textbf{0.0013(-63.9\%)}&\textbf{0.0010(-56.5\%)}&\textbf{0.0009(-52.6\%)}&\textbf{0.0009(-43.8\%)} \\
\hline
\end{tabular}
}
\end{table}

\vspace{-20pt}
\subsection{Scaling}
\begin{figure}[ht]
\centering
\subfigure[]{
\label{dim}
\scalebox{0.9}{
\includegraphics[width=0.45\linewidth]{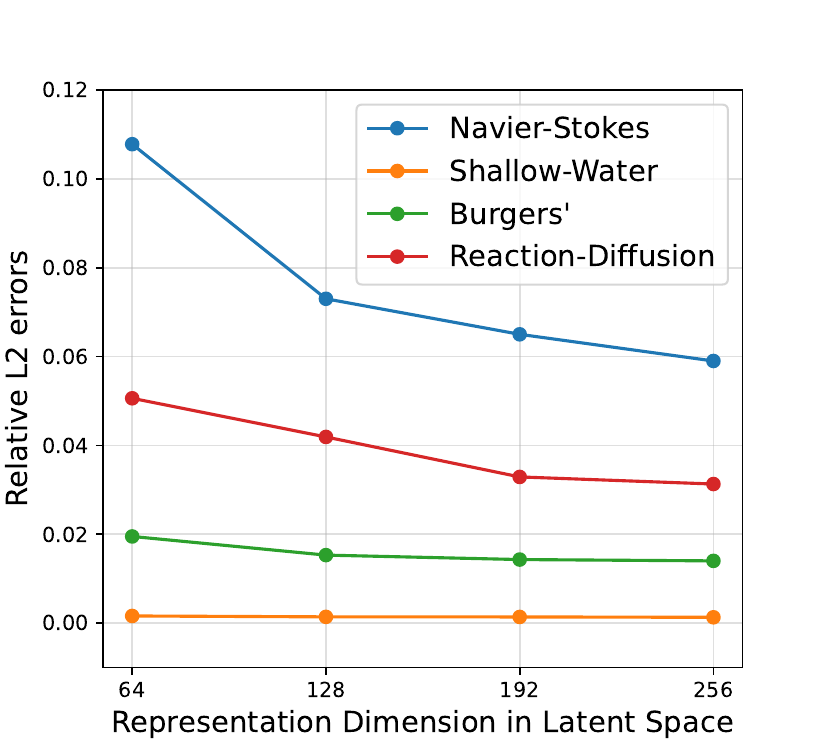}
}
}
\noindent
\subfigure[]{
\label{num}
\scalebox{0.9}{
\includegraphics[width=0.45\linewidth]{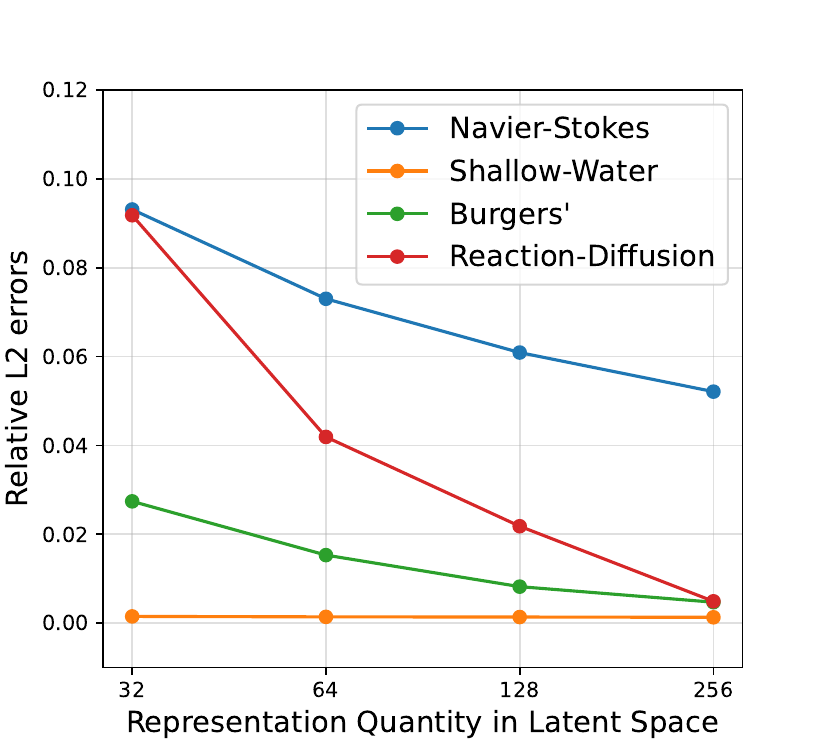}	
}
}
\vspace{-10pt}
\caption{Results of scaling experiments. (a) Impact of representation token dimension on solution precision of various PDE problems. (b) Impact of representation token quantity on solution precision of various PDE problems.}
\label{scaling}
\vspace{-10pt}
\end{figure}

We conduct scaling experiments to demonstrate the solution precision on all time-dependent PDE problems of our proposed LNOP framework as the number and dimension of representation tokens vary. The results in Figure \ref{dim} indicate the solution precision of the Navier-Stokes equation consistently improves as the token dimension increases from $32$ to as large as $256$, while that of the other three PDEs gradually saturates when the token dimension reaches $192$. The results in Figure \ref{num} show that, aside from the shallow-water equation which consistently maintains precise solution, the solution precision of the other three PDEs improves continuously with an increasing number of representation tokens.

\begin{table}[ht]
\centering
\caption{The solution precision of LNOP pretrained with different approach or finetuned with different component on various PDE problems respectively. Relative L2 is recorded. The best result is in bold.}
\label{tab-ab}
\scalebox{0.85}{
\begin{tabular}{|c|c|c|c|c|}
\hline
Model&Navier-Stokes&Shallow-Water&Burgers'&Reaction-Diffusion\\
\hline
LNOP-S(pretrain)&0.0730&0.0014&0.0153&0.0419\\
LNOP-S(finetune-All)&\textbf{0.0456}&\textbf{0.0004}&\textbf{0.0112}&\textbf{0.0236}\\
LNOP-S(finetune-PhCA)&0.0722&0.0014&0.0151&0.0411\\
LNOP-S(finetune-Others)&0.0526&0.0010&0.0117&0.0263\\
LNOP-S(two-stage)&0.2236&0.0122&0.0817&0.1274\\
\hline
\end{tabular}
}
\vspace{-15pt}
\end{table}

\vspace{-10pt}
\subsection{Ablation Study}
\vspace{-1pt}
\label{ablation}
We conduct ablation study to investigate the impact of finetuning different components in the LNOP framework on the solution precision. Specifically, we compare the following three finetuning scenarios: i) all parameters; ii) only the PhCA encoder/decoder; iii) only the components other than the PhCA encoder/decoder, including the input and output projector and the propagator. The results in Table \ref{tab-ab} show that, although finetuning all parameters achieves the highest solution precision, finetuning the components other than the PhCA encoder/decoder can yield higher precision than finetuning only the PhCA encoder/decoder. This indicates that the PhCA encoder/decoder effectively learn universal transformation for representation extraction from multiple physical systems.

We also try modifying the LNOP framework into two-stage approach. In the first stage, we pretrain a PhCA-based autoencoder on the hybrid dataset using reconstruction task. The autoencoder takes several frames of time-dependent PDEs as input, extracts representations in the latent space, and reconstructs them back into PDE information. In the second stage, we train propagators using autoregressive task on each single PDE problem to predict the temporal evolution. The propagator iterates the PDE representations from the initial state to the final state in the latent space. This implies that the PhCA encoder/decoder is only used at the initial and final moments of the PDE system, with the intermediate temporal evolution prediction relying solely on the propagator.

The result in the last row of Table \ref{tab-ab} indicates that the two-stage LNOP approach performs not as well as the end-to-end one. This may be due to the discrepancy between the representation spaces required by the reconstruction task and PDE solution task, which introduces a mismatch between the autoencoder and the propagator.

\section{Conclusion}
\label{conclusion}
We propose the Latent Neural Operator Pretraining (LNOP) framework to learn universal transformation for extracting representations of different PDEs in a shared latent space across multiple physical systems. We pretrains the LNO backbone on hybrid dataset comprising multiple time-dependent PDE problems and compare its solution precision under different finetuning conditions for both in-distribution and out-of-distribution time-dependent PDE problems. Through a series of experiments, we verifies the precision improvement gained from learning shared representations through pretraining, and also validates the transfer capability and data efficiency brought by the learned universal transformation. 

Our work also has some limitations. First, our purely data-driven method does not leverage prior knowledge from different PDEs, which may compromise the solution precision. Additionally, our method does not completely separate PDE representation learning from PDE time evolution prediction, which slows down the pretraining process. 

Future work should focus on how to incorporate physical prior knowledge as constraints or additional modalities into the PDE solving process, how to improve PDE representation learning, and how to achieve PDE time evolution estimation entirely in the latent space.

%
%
%
\bibliographystyle{splncs04}
\bibliography{0590}
\end{document}